\def\year{2019}\relax
\documentclass[letterpaper]{article} 
\usepackage{aaai19}  
\usepackage{times}  
\usepackage{helvet}  
\usepackage{courier}  
\usepackage{url}  
\usepackage{graphicx}  
\usepackage{multirow}
\usepackage{epsfig}
\usepackage{amsmath}
\usepackage{amssymb,amsfonts}
\usepackage{bm}
\usepackage{CJK}
\usepackage{subcaption}

\begin{document}
%
\nocopyright

\newcommand*{\affaddr}[1]{#1} 
\newcommand*{\affmark}[1][*]{\textsuperscript{#1}}
\newcommand*{\email}[1]{\texttt{#1}}

\title{Unsupervised Machine Commenting with Neural Variational Topic Model}
\author{Shuming Ma\affmark[1]\thanks{Contribution during internship at Microsoft Research
Asia.}, Lei Cui\affmark[2], Furu Wei\affmark[2], Xu Sun\affmark[1]\\
\affaddr{\affmark[1]MOE Key Lab of Computational Linguistics, School of EECS, Peking University}\\
\affaddr{\affmark[2]Microsoft Research Asia}\\
\email{\{shumingma,xusun\}@pku.edu.cn}\\
\email{\{lecu,fuwei\}@microsoft.com}
}

\maketitle
\begin{abstract}
Article comments can provide supplementary opinions and facts for readers, thereby increase the attraction and engagement of articles. Therefore, automatically commenting is helpful in improving the activeness of the community, such as online forums and news websites. Previous work shows that training an automatic commenting system requires large parallel corpora. Although part of articles are naturally paired with the comments on some websites, most articles and comments are unpaired on the Internet. To fully exploit the unpaired data, we completely remove the need for parallel data and propose a novel unsupervised approach to train an automatic article commenting model, relying on nothing but unpaired articles and comments. Our model is based on a retrieval-based commenting framework, which uses news to retrieve comments based on the similarity of their topics. The topic representation is obtained from a neural variational topic model, which is trained in an unsupervised manner. We evaluate our model on a news comment dataset. Experiments show that our proposed topic-based approach significantly outperforms previous lexicon-based models. The model also profits from paired corpora and achieves state-of-the-art performance under semi-supervised scenarios.
\end{abstract}

\section{Introduction}

Making article comments is a fundamental ability for an intelligent machine to understand the article and interact with humans.
It provides more challenges because commenting requires the abilities of comprehending the article, summarizing the main ideas, mining the opinions, and generating the natural language. Therefore, machine commenting is an important problem faced in building an intelligent and interactive agent. Machine commenting is also useful in improving the activeness of communities, including online forums and news websites. Article comments can provide extended information and external opinions for the readers to have a more comprehensive understanding of the article. Therefore, an article with more informative and interesting comments will attract more attention from readers. Moreover, machine commenting can kick off the discussion about an article or a topic, which helps increase user engagement and interaction between the readers and authors.

\begin{table*}[t]
\centering
    \begin{tabular}{p{17cm}}
    \hline
    \textbf{Title:}  LeBron James, Lakers to face Warriors on Christmas Day \\
    \hline
    \textbf{Body:} LeBron James and the Los Angeles Lakers will face the defending champion Golden State Warriors on Christmas Day, league sources confirmed. James and the Lakers' game at Golden State will be one of several nationally televised NBA games on the holiday. Joel Embiid, Ben Simmons and the Philadelphia 76ers will visit Kyrie Irving, Jayson Tatum and the Boston Celtics, sources confirmed. The Utah Jazz will host the Portland Trail Blazers in another game, ESPN's Chris Haynes reports. The New York Knicks will host the Milwaukee Bucks, sources confirmed. And the Oklahoma City Thunder will visit the Houston Rockets, ESPN's Tim MacMahon reports. James' first Christmas Day game as a Laker will be the day's marquee matchup. James, who signed a four-year, \$153 million contract with the Lakers in July, faced the Warriors in each of the past four NBA Finals as a member of the Cleveland Cavaliers. The Sixers and Celtics -- two teams that boast some of the top young talent in the NBA -- will face off in a rematch of their Eastern Conference semifinals series, which the Celtics won in five games. The Bucks will play on the holiday for the first time since 1977. It is unclear whether Knicks All-Star Kristaps Porzingis will be available for the Christmas Day game. He is currently rehabbing a torn ACL. Porzingis suffered the ACL tear in an early February game against the Bucks. The Lakers-Warriors, Celtics-Sixers and Knicks-Bucks Christmas Day games were first reported by The New York Times.
Information from ESPN's Ian Begley was used in this report.\\
    \hline
    \textbf{Comment 1:} No finals rematch between Cavs and warriors?? \\
    \textbf{Comment 2:} Adam Silver is straight up saying, "It was always LeBron vs Warriors." \\
    \textbf{Comment 3:} LeBron has missed Christmas Day for 15 years now... \\
    \textbf{Comment 4:} this is so dumb honestly, should've been lakers v celtics \\
    \textbf{Comment 5:} whyyyyyy. Warriors vs Rockets would be so much more competitive. Let the Lakers play the Celtics, Sixers could play Thunder, and Raps vs Spurs instead of Jazz v Blazers. This seems obvious no? \\
    \hline
    \end{tabular}
    \caption{An example of an article and five selected comments.}
    \label{intro_example}
\end{table*}

Because of the advantage and importance described above, more recent studies have focused on building a machine commenting system with neural models~\cite{QinEA2018}. One bottleneck of neural machine commenting models is the requirement of a large parallel dataset. However, the naturally paired commenting dataset is loosely paired. Qin et al.~\shortcite{QinEA2018} were the first to propose the article commenting task and an article-comment dataset. The dataset is crawled from a news website, and they sample 1,610 article-comment pairs to annotate the relevance score between articles and comments. The relevance score ranges from 1 to 5, and we find that only 6.8\% of the pairs have an average score greater than 4. It indicates that the naturally paired article-comment dataset contains a lot of loose pairs, which is a potential harm to the supervised models. Besides, most articles and comments are unpaired on the Internet. For example, a lot of articles do not have the corresponding comments on the news websites, and the comments regarding the news are more likely to appear on social media like Twitter. Since comments on social media are more various and recent, it is important to exploit these unpaired data.

Another issue is that there is a semantic gap between articles and comments. In machine translation and text summarization, the target output mainly shares the same points with the source input. However, in article commenting, the comment does not always tell the same thing as the corresponding article. Table~\ref{intro_example} shows an example of an article and several corresponding comments. The comments do not directly tell what happened in the news, but talk about the underlying topics (e.g. NBA Christmas Day games, LeBron James). However, existing methods for machine commenting do not model the topics of articles, which is a potential harm to the generated comments.

To this end, we propose an unsupervised neural topic model to address both problems. For the first problem, we completely remove the need of parallel data and propose a novel unsupervised approach to train a machine commenting system, relying on nothing but unpaired articles and comments. For the second issue, we bridge the articles and comments with their topics. Our model is based on a retrieval-based commenting framework, which uses the news as the query to retrieve the comments by the similarity of their topics. The topic is represented with a variational topic, which is trained in an unsupervised manner.

The contributions of this work are as follows:
\begin{itemize}
\item To the best of our knowledge, we are the first to explore an unsupervised learning approach for machine commenting. We believe our exploration can shed some light on how to exploit unpaired data for a follow-up study on machine commenting.
\item We propose using the topics to bridge the semantic gap between the articles and the comments. We introduce a variation topic model to represent the topics, and match the articles and the comments by the similarity of their topics. We evaluate our model on a news comment dataset. Experiments show that our topic-based approach significantly outperforms previous lexical-based models.
\item We explore semi-supervised scenarios, and experimental results show that our model achieves better performance than previous supervised models.
\end{itemize}

\section{Machine Commenting}

In this section, we highlight the research challenges of machine commenting, and provide some solutions to deal with these challenges.

\subsection{Challenges}\label{challenge}

Here, we first introduce the challenges of building a well-performed machine commenting system.

\subsubsection{Mode Collapse Problem}

The generative model, such as the popular sequence-to-sequence model, is a direct choice for supervised machine commenting. One can use the title or the content of the article as the encoder input, and the comments as the decoder output. However, we find that the mode collapse problem is severed with the sequence-to-sequence model. Despite the input articles being various, the outputs of the model are very similar. The reason mainly comes from the contradiction between the complex pattern of generating comments and the limited parallel data.
In other natural language generation tasks, such as machine translation and text summarization, 
the target output of these tasks is strongly related to the input, and most of the required information is involved in the input text. However, the comments are often weakly related to the input articles, and part of the information in the comments is external. Therefore, it requires much more paired data for the supervised model to alleviate the mode collapse problem.

\subsubsection{Falsely Negative Samples}

One article can have multiple correct comments, and these comments can be very semantically different from each other. However, in the training set, there is only a part of the correct comments, so the other correct comments will be falsely regarded as the negative samples by the supervised model. Therefore, many interesting and informative comments will be discouraged or neglected, because they are not paired with the articles in the training set.

\subsubsection{Semantic Gap}

There is a semantic gap between articles and comments. In machine translation and text summarization, the target output mainly shares the same points with the source input. However, in article commenting, the comments often have some external information, or even tell an opposite opinion from the articles. Therefore, it is difficult to automatically mine the relationship between articles and comments.




\subsection{Solutions}

Facing the above challenges, we provide three solutions to the problems.

\subsubsection{Retrieval Model}

Given a large set of candidate comments, the retrieval model can select some comments by matching articles with comments. Compared with the generative model, the retrieval model can achieve more promising performance. First, the retrieval model is less likely to suffer from the mode collapse problem. Second, the generated comments are more predictable and controllable (by changing the candidate set). Third, the retrieval model can be combined with the generative model to produce new comments (by adding the outputs of generative models to the candidate set).

\subsubsection{Unsupervised Learning}

The unsupervised learning method is also important for machine commenting to alleviate the problems descried above. Unsupervised learning allows the model to exploit more data, which helps the model to learn more complex patterns of commenting and improves the generalization of the model. Many comments provide some unique opinions, but they do not have paired articles. For example, many interesting comments on social media (e.g. Twitter) are about recent news, but require redundant work to match these comments with the corresponding news articles. With the help of the unsupervised learning method, the model can also learn to generate these interesting comments. Additionally, the unsupervised learning method does not require negative samples in the training stage, so that it can alleviate the negative sampling bias.




\subsubsection{Modeling Topic}

Although there is semantic gap between the articles and the comments, we find that most articles and comments share the same topics. Therefore, it is possible to bridge the semantic gap by modeling the topics of both articles and comments. It is also similar to how humans generate comments. Humans do not need to go through the whole article but are capable of making a comment after capturing the general topics.




\section{Proposed Approach}

We now introduce our proposed approach as an implementation of the solutions above. We first give the definition and the denotation of the problem. Then, we introduce the retrieval-based commenting framework. After that, a neural variational topic model is introduced to model the topics of the comments and the articles. Finally, semi-supervised training is used to combine the advantage of both supervised and unsupervised learning.

\subsection{Retrieval-based Commenting}

Given an article, the retrieval-based method aims to retrieve a comment from a large pool of candidate comments. The article consists of a title $\bm{t}$ and a body $\bm{b}$. The comment pool is formed from a large scale of candidate comments $[c_1,c_2,\cdots,c_N]$, where $N$ is the number of the unique comments in the pool. In this work, we have 4.5 million human comments in the candidate set, and the comments are various, covering different topics from pets to sports.

The retrieval-based model should score the matching between the upcoming article and each comments, and return the comments which is matched with the articles the most. Therefore, there are two main challenges in retrieval-based commenting. One is how to evaluate the matching of the articles and comments. The other is how to efficiently compute the matching scores because the number of comments in the pool is large. 

To address both problems, we select the ``dot-product'' operation to compute matching scores. More specifically, the model first computes the representations of the article $\bm{h_a}$ and the comments $\bm{h_c}$. Then the score between article $\bm{a}$ and comment $\bm{c}$ is computed with the ``dot-product'' operation:
\begin{equation}
s(\bm{a}, \bm{c})=\bm{h_a}^T\bm{h_c}
\end{equation}

The dot-product scoring method has proven a successful in a matching model~\cite{smartreply}. The problem of finding datapoints with the largest dot-product values is called Maximum Inner Product Search (MIPS), and there are lots of solutions to improve the efficiency of solving this problem. Therefore, even when the number of candidate comments is very large, the model can still find comments with the highest efficiency. However, the study of the MIPS is out of the discussion in this work. We refer the readers to relevant articles for more details about the MIPS~\cite{mips4,mips1,mips3,mips2}. Another advantage of the dot-product scoring method is that it does not require any extra parameters, so it is more suitable as a part of the unsupervised model.

\subsection{Neural Variational Topic Model}

We obtain the representations of articles $\bm{h_a}$ and comments $\bm{h_c}$ with a neural variational topic model. The neural variational topic model is based on the variational autoencoder framework,  so it can be trained in an unsupervised manner. The model encodes the source text into a representation, from which it reconstructs the text.

We concatenate the title and the body to represent the article. In our model, the representations of the article and the comment are obtained in the same way. For simplicity, we denote both the article and the comment as ``document''. Since the articles are often very long (more than 200 words), we represent the documents into bag-of-words, for saving both the time and memory cost. We denote the bag-of-words representation as $\bm{X} \in \mathbb{R}^{|V|}$, where $x_i \in \mathbb{R}^{|V|}$ is the one-hot representation of the word at $i^{th}$ position, and $|V|$ is the number of words in the vocabulary. The encoder $q(\bm{h}|\bm{X})$ compresses the bag-of-words representations $\bm{X}$ into topic representations $\bm{h} \in \mathbb{R}^{K}$:
\begin{equation}
z=\tanh{(W_1X+b_1)}
\end{equation}
\begin{equation}
q(h|X)=\tanh{(W_2z+b_2)}
\end{equation}
where $W_1$, $W_2$, $b_1$, and $b_2$ are the trainable parameters. Then the decoder $p(\bm{X}|\bm{h})$ reconstructs the documents by independently generating each words in the bag-of-words:
\begin{equation}
p(\bm{X}|\bm{h})=\prod_{i=1}^{N}p(\bm{x}_i|\bm{h})
\end{equation}
\begin{equation}
p(\bm{x}_i|\bm{h})=softmax(\bm{h}^{T}\bm{M}\bm{x}_i)
\end{equation}
where $N$ is the number of words in the bag-of-words, and $\bm{M} \in \mathbb{R}^{K \times |V|}$ is a trainable matrix to map the topic representation into the word distribution.

In order to model the topic information, we use a Dirichlet prior rather than the standard Gaussian prior. However, it is difficult to develop an effective reparameterization function for the Dirichlet prior to train VAE. Therefore, following~\cite{nvlda}, we use the Laplace approximation~\cite{laplace} to Dirichlet prior $p(\bm{h})=\mathcal{LN}(\bm{h}|\bm{\mu_0},\bm{\sigma_0})$:
\begin{equation}
\mu_{0i}=\log{\alpha_i}-\frac{1}{K}\sum_{j=1}^{K}\log{\alpha_{j}} 
\end{equation}
\begin{equation}
\sigma_{0i}=\frac{1}{\alpha_i}{(1-\frac{2}{K})}+\frac{1}{K^2}{\sum_{j=1}^{K}{\frac{1}{\alpha_j}}}
\end{equation}
where $\mathcal{LN}$ denotes the logistic normal distribution, $K$ is the number of topics, and $\bm{\alpha}$ is a parameter vector. Then, the variational lower bound is written as:
\begin{equation}\label{loss}
\begin{split}
\mathcal{L}= &-\frac{1}{2}\Big(\bm{\sigma}^{T}\bm{\sigma}_{0}^{-1}+(\bm{\mu}_0-\bm{\mu})^{T}\text{diag}(\bm{\sigma}_{0}^{-1})(\bm{\mu}_0-\bm{\mu})\\
 &-K+\log{\frac{|\bm{\sigma}|}{|\bm{\sigma}_0|}}\Big) + \sum_{i=1}^{N}\log{p(\bm{x}_i|\bm{\theta})}
\end{split}
\end{equation}
where the first term is the KL-divergence loss and the second term is the reconstruction loss. The mean $\bm{\mu}$ and the variance $\bm{\sigma}$ are computed as follows:
\begin{equation}
\bm{\mu}=W_{3}h+b_{3}
\end{equation}
\begin{equation}
\bm{\sigma}=W_{4}h+b_{4}
\end{equation}
We use the $\bm{\mu}$ and $\bm{\sigma}$ to generate the samples $\theta=\mu+\sigma^{1/2}\epsilon$ by sampling $\epsilon \sim \mathcal{N}(0,\bm{1})$, from which we reconstruct the input $\bm{X}$.

At the training stage, we train the neural variational topic model with the Eq.~\ref{loss}. At the testing stage, we use $q(\bm{h}|\bm{X})$ to compute the topic representations of the article $\bm{h}_a$ and the comment $\bm{h}_c$.

\subsection{Training}

In addition to the unsupervised training, we explore a semi-supervised training framework to combine the proposed unsupervised model and the supervised model. In this scenario we have a paired dataset that contains article-comment parallel contents $(\bm{s}, \bm{c}) \in \mathbb{L}$, and an unpaired dataset that contains the documents (articles or comments) $\bm{d} \in \mathbb{U}$. The supervised model is trained on $\mathbb{L}$ so that we can learn the matching or mapping between articles and comments. By sharing the encoder of the supervised model and the unsupervised model, we can jointly train both the models with a joint objective function:
\begin{equation}\label{total}
\mathcal{L}=\mathcal{L}_{unsuper}+\lambda \mathcal{L}_{super}
\end{equation}
where $\mathcal{L}_{unsuper}$ is the loss function of the unsupervised learning (Eq.~ref{loss}), $\mathcal{L}_{super}$ is the loss function of the supervised learning (e.g. the cross-entropy loss of Seq2Seq model), and $\lambda$ is a hyper-parameter to balance two parts of the loss function.
Hence, the model is trained on both unpaired data $\mathbb{U}$, and paired data $\mathbb{L}$.

\section{Experiments}

\subsection{Datasets}

We select a large-scale Chinese dataset~\cite{QinEA2018} with millions of real comments and a human-annotated test set to evaluate our model.
The dataset is collected from Tencent News\footnote{\url{news.qq.com}}, which is one of the most popular Chinese websites for news and opinion articles. The dataset consists of 198,112 news articles. Each piece of news contains a title, the content of the article, and a list of the users' comments. Following the previous work~\cite{QinEA2018}, we tokenize all text with the popular
python package Jieba\footnote{https://github.com/fxsjy/jieba}, and filter out short articles with less than 30 words in content and those with less than 20 comments. The dataset is split into training/validation/test sets, and they contain 191,502/5,000/1,610 pieces of news, respectively. The whole dataset has a vocabulary size of 1,858,452. The average lengths of the article titles and content are 15 and 554 Chinese words. The average comment length is 17 words.

\subsection{Implementation Details}
The hidden size of the model is 512, and the batch size is 64. The number of topics $K$ is 100. The weight $\lambda$ in Eq.~\ref{total} is 1.0 under the semi-supervised setting. We prune the vocabulary, and only leave 30,000 most frequent words in the vocabulary. We train the model for 20 epochs with the Adam optimizing algorithms~\cite{KingmaBa2014}. In order to alleviate the KL vanishing problem, we set the initial learning to $5^{-5}$, and use batch normalization~\cite{batchnorm} in each layer. We also gradually increase the KL term from 0 to 1 after each epoch. 

\begin{table*}[t]
	\centering
	\begin{tabular}{c | l | c c | c c c c c}
		\hline
         &
		\multicolumn{1}{c}{\textbf{Model}} &
        \multicolumn{1}{|c}{\textbf{Paired}} &
       \multicolumn{1}{c|}{\textbf{Unpaired}} &
		\multicolumn{1}{c}{\textbf{Recall@1}} & 
		\multicolumn{1}{c}{\textbf{Recall@5}} &  
		\multicolumn{1}{c}{\textbf{Recall@10}} &
        \multicolumn{1}{c}{\textbf{MR}} &
        \multicolumn{1}{c}{\textbf{MRR}} \\ \hline
        \multirow{4}{*}{Unsupervised} & TF-IDF & - & 4.8M & 3.41 & 7.51 & 10.62 & 19.06 & 0.0095 \\
        & NVDM & - & 4.8M & 12.73 & 53.16 & 74.47 & 7.62 & 0.3053 \\ 
        & LDA & - & 4.8M & 17.14 & 56.39 & 74.65 & 7.69 & 0.3512 \\ 
        & Proposed & - & 4.8M & \textbf{22.48} & \textbf{67.45} & \textbf{86.15} & \textbf{5.35} & \textbf{0.4186} \\ \hline 
        \multirow{4}{*}{Supervised} & S2S$_{1}$ & 50K & - & 7.20 & 40.43 & 67.14 & 9.39 & 0.2335 \\
        & S2S$_{2}$ & 4.8M & - & 10.68 & 47.39 & 72.60 & 8.34 & 0.2787 \\ 
        & IR$_{1}$ & 50K & - & 35.34 & 79.01 & 92.92 & 3.95 & 0.5384 \\
        & IR$_{2}$ & 4.8M & - & \textbf{45.83} & \textbf{88.19} & \textbf{94.02} & \textbf{3.57} & \textbf{0.6375} \\ \hline
        \multirow{4}{*}{Semi-supervised} & Proposed+S2S$_{1}$ & 50K & 4.8M & 11.73 & 50.31 & 75.46 & 7.86 & 0.2930 \\
        & Proposed+S2S$_{2}$ & 4.8M & 4.8M & 14.61 & 52.76 & 77.22 & 6.32 & 0.3175 \\ 
        & Proposed+IR$_{1}$ & 50K & 4.8M & 43.85 & 84.96 & 93.29 & 3.45 & 0.6102 \\ 
        & Proposed+IR$_{2}$ & 4.8M & 4.8M & \textbf{53.91} & \textbf{86.77} & \textbf{94.66} & \textbf{3.02} & \textbf{0.6822} \\ \hline
        
	\end{tabular}
	\caption{The performance of the unsupervised models and supervised models under the retrieval evaluation settings. (Recall@k, MRR: higher is better; MR: lower is better.)}\label{tab_ret}
\end{table*}

\begin{table*}[t]
	\centering
	\begin{tabular}{c | l | c c | c c c c}
		\hline
        &
		\multicolumn{1}{c}{\textbf{Model}} &
        \multicolumn{1}{|c}{\textbf{Paired}} &
       \multicolumn{1}{c|}{\textbf{Unpaired}} &
		\multicolumn{1}{c}{\textbf{METEOR}} & 
		\multicolumn{1}{c}{\textbf{ROUGE}} &  
		\multicolumn{1}{c}{\textbf{CIDEr}} &
        \multicolumn{1}{c}{\textbf{BLEU}} \\ \hline
        \multirow{4}{*}{Unsupervised} & TF-IDF & - & 4.8M & 0.005 & 0.124 & 0.016 & 0.197  \\
        & NVDM & - & 4.8M & 0.101 & 0.155 & 0.018 & 0.250  \\ 
        & LDA & - & 4.8M & 0.085 & 0.148 & 0.017 & 0.248 \\ 
        & Proposed & - & 4.8M & \textbf{0.110} & \textbf{0.162} & \textbf{0.022} & \textbf{0.261}  \\ \hline
        \multirow{4}{*}{Supervised} 
        & S2S$_{1}$ & 50K & - & 0.029 & 0.093 & 0.001 & 0.078 \\
        & S2S$_{2}$ & 4.8M & - & 0.031 & 0.099 & 0.004 & 0.104 \\ 
        & IR$_{1}$ & 50K & - & 0.113 & 0.162 & 0.021 & 0.261 \\
        & IR$_{2}$ & 4.8M & - & \textbf{0.115} & \textbf{0.167} & \textbf{0.032} & \textbf{0.283}\\ \hline
        \multirow{4}{*}{Semi-supervised} & Proposed+S2S$_1$ & 50K & 4.8M & 0.041 & 0.104 & 0.002 & 0.100\\
        & Proposed+S2S$_2$ & 4.8M & 4.8M & 0.049 & 0.109 & 0.005 & 0.112\\ 
        & Proposed+IR$_1$ & 50K & 4.8M & 0.122 & 0.176 & 0.030 & 0.275 \\ 
        & Proposed+IR$_2$ & 4.8M & 4.8M & \textbf{0.130} & \textbf{0.187} & \textbf{0.041} & \textbf{0.294} \\ \hline
		
	\end{tabular}
	\caption{The performance of the unsupervised models and supervised models under the generative evaluation settings. (METEOR, ROUGE, CIDEr, BLEU: higher is better.)}\label{tab_gen}
\end{table*}

\subsection{Baselines}

We compare our model with several unsupervised models and supervised models.

\noindent Unsupervised baseline models are as follows:

\begin{itemize}
\item \textbf{TF-IDF (Lexical, Non-Neural)} is an important unsupervised baseline. We use the concatenation of the title and the body as the query to retrieve the candidate comment set by means of the similarity of the tf-idf value. The model is trained on unpaired articles and comments, which is the same as our proposed model.
\item \textbf{LDA (Topic, Non-Neural)} is a popular unsupervised topic model, which discovers the abstract "topics" that occur in a collection of documents. We train the LDA with the articles and comments in the training set. The model retrieves the comments by the similarity of the topic representations.
\item \textbf{NVDM (Lexical, Neural)} is a VAE-based approach for document modeling~\cite{NVDM}. We compare our model with this baseline to demonstrate the effect of modeling topic. 
\end{itemize}

\noindent The supervised baseline models are:
\begin{itemize}
\item \textbf{S2S (Generative)}~\cite{seq2seq} is a supervised generative model based on the sequence-to-sequence network with the attention mechanism~\cite{attention}. The model uses the titles and the bodies of the articles as the encoder input, and generates the comments with the decoder.
\item \textbf{IR (Retrieval)}~\cite{QinEA2018} is a supervised retrieval-based model, which trains a convolutional neural network (CNN) to take the articles and a comment as inputs, and output the relevance score. The positive instances for training are the pairs in the training set, and the negative instances are randomly sampled using the negative sampling technique~\cite{MikolovEA2013}.
\end{itemize}

\begin{figure}[t]
	\centering
	\includegraphics[width=0.9\linewidth]{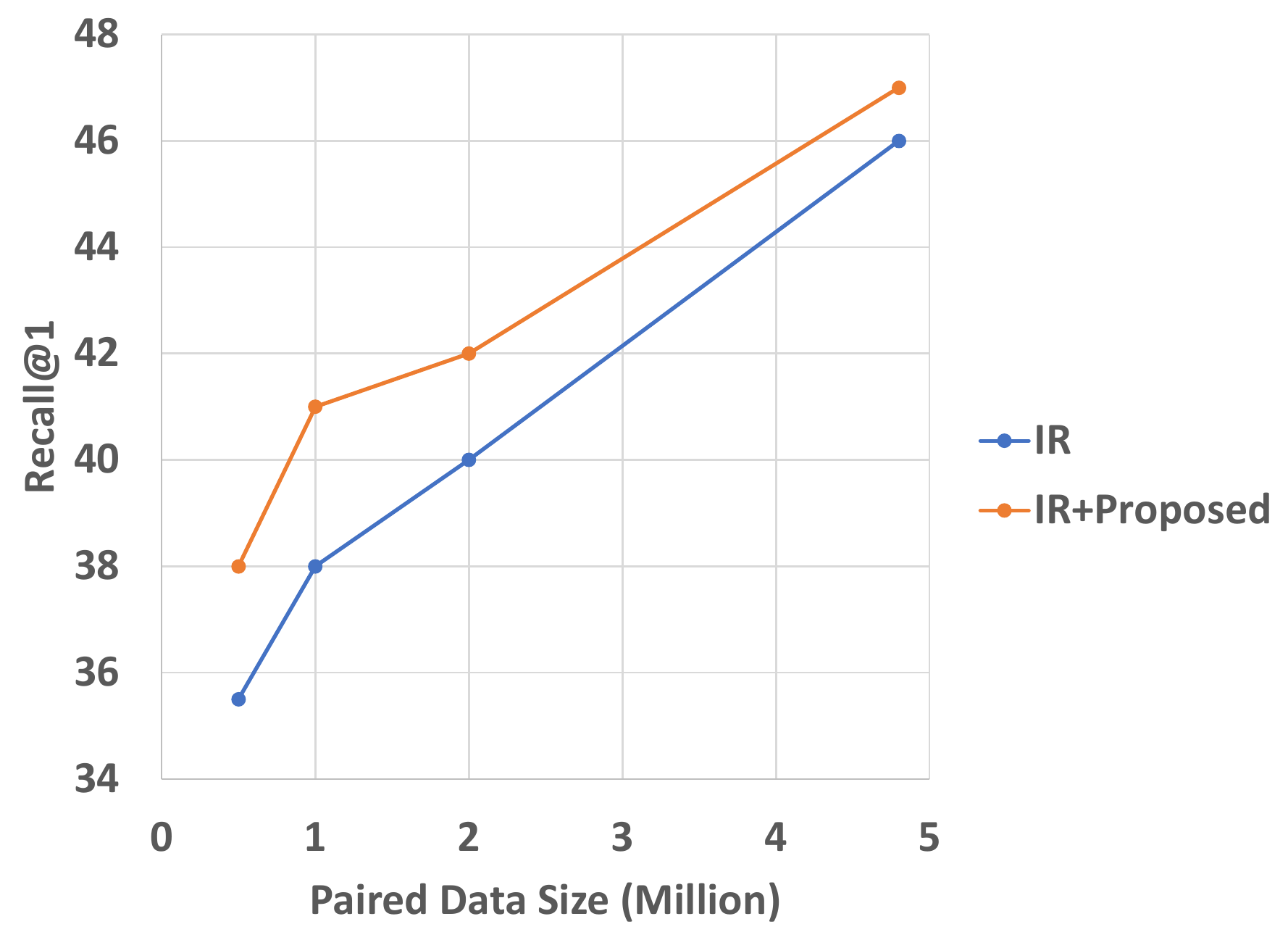}
	\caption{The performance of the supervised model and the semi-supervised model trained on different paired data size.}\label{semi_fig}
\end{figure}

\begin{figure}[t]
	\centering
	\subcaptionbox{TF-IDF}{\includegraphics[width=0.44\linewidth]{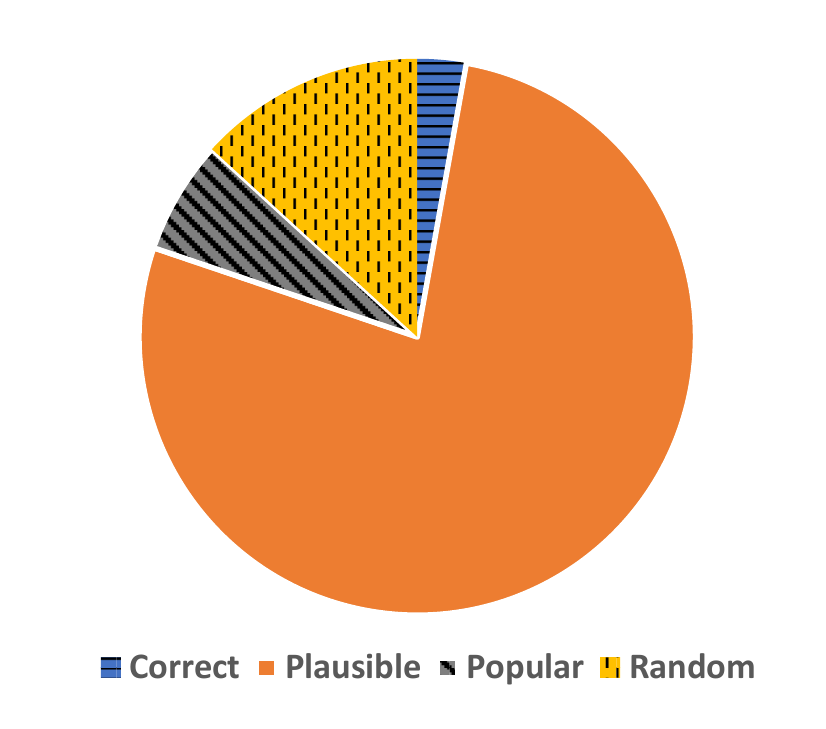}}
	\subcaptionbox{S2S}{\includegraphics[width=0.44\linewidth]{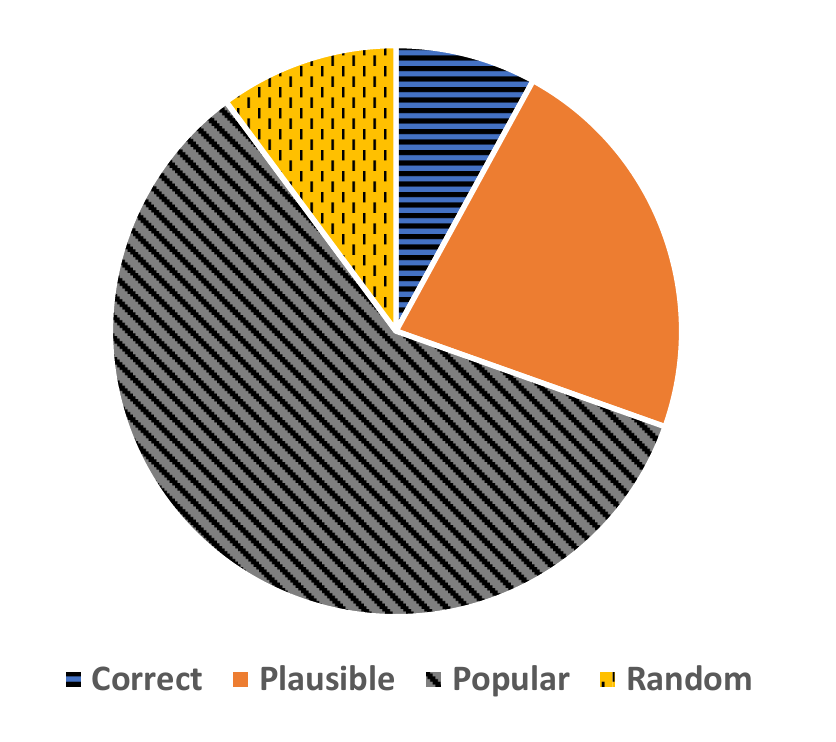}}\\
    \subcaptionbox{IR}{\includegraphics[width=0.44\linewidth]{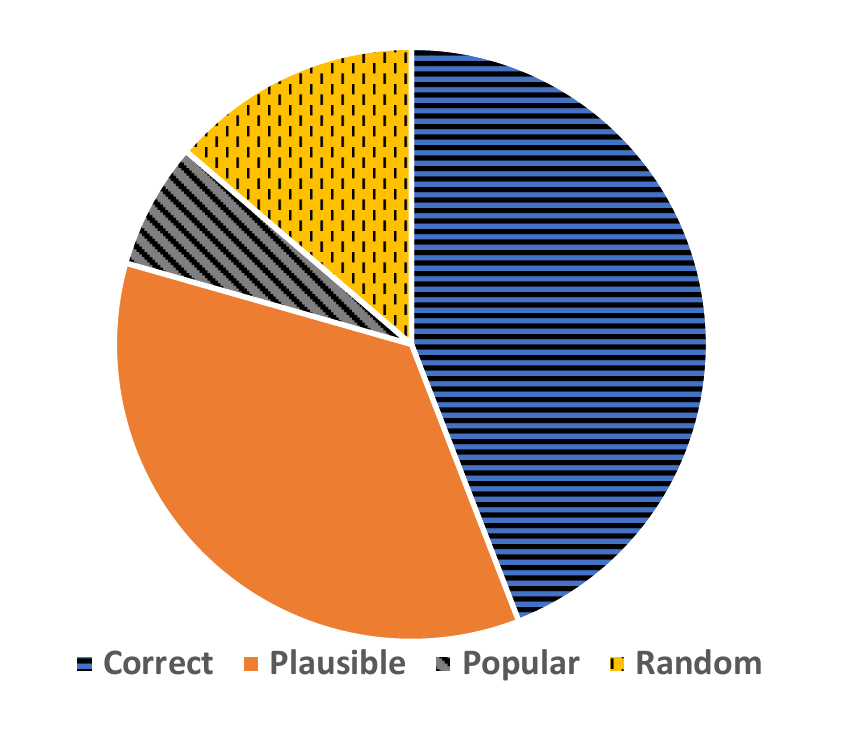}}
	\subcaptionbox{Proposed+IR}{\includegraphics[width=0.44\linewidth]{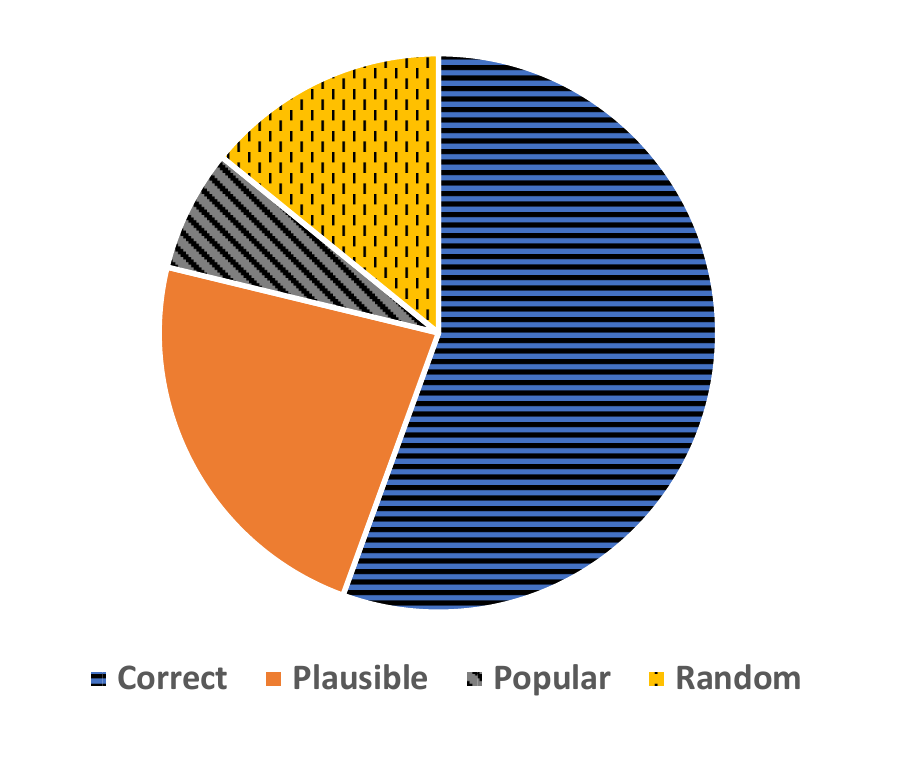}}
	\caption{Error types of comments generated by different models.}\label{error_fig}
\end{figure}

\subsection{Retrieval Evaluation}

For text generation, automatically evaluate the quality of the generated text is an open problem. In particular, the comment of a piece of news can be various, so it is intractable to find out all the possible references to be compared with the model outputs. 

Inspired by the evaluation methods of dialogue models, we formulate the evaluation as a ranking problem. Given a piece of news and a set of candidate comments, the comment model should return the rank of the candidate comments. The candidate comment set consists of the following parts:

\textbf{Correct:} The ground-truth comments of the corresponding news provided by the human.

\textbf{Plausible}: The 50 most similar comments to the news. We use the news as the query to retrieve the comments that appear in the training set based on the cosine similarity of their tf-idf values. We select the top 50 comments that are not the correct comments as the plausible comments.

\textbf{Popular:} The 50 most popular comments from the dataset. We count the frequency of each comments in the training set, and select the 50 most frequent comments to form the popular comment set. The popular comments are the general and meaningless comments, such as ``Yes'', ``Great'', ``That's right', and ``Make Sense''. These comments are dull and do not carry any information, so they are regarded as incorrect comments.

\textbf{Random:} After selecting the correct, plausible, and popular comments, we fill the candidate set with randomly selected comments from the training set so that there are 200 unique comments in the candidate set.

Following previous work, we measure the rank in terms of the following metrics:

\textbf{Recall@k:} The proportion of human comments found in the top-k recommendations.

\textbf{Mean Rank (MR):} The mean rank of the human comments.

\textbf{Mean Reciprocal Rank (MRR):} The mean reciprocal rank of the human comments.

The evaluation protocol is compatible with both retrieval models and generative models. The retrieval model can directly rank the comments by assigning a score for each comment, while the generative model can rank the candidates by the model's log-likelihood score.

\subsubsection{Results} 

Table~\ref{tab_ret} shows the performance of our models and the baselines in retrieval evaluation. We first compare our proposed model with other popular unsupervised methods, including TF-IDF, LDA, and NVDM. TF-IDF retrieves the comments by similarity of words rather than the semantic meaning, so it achieves low scores on all the retrieval metrics. The neural variational document model is based on the neural VAE framework. It can capture the semantic information, so it has better performance than the TF-IDF model. LDA models the topic information, and captures the deeper relationship between the article and comments, so it achieves improvement in all relevance metrics. Finally, our proposed model outperforms all  these unsupervised methods, mainly because the proposed model learns both the semantics and the topic information.

We also evaluate two popular supervised models, i.e. seq2seq and IR. Since the articles are very long, we find either RNN-based or CNN-based encoders cannot hold all the words in the articles, so it requires limiting the length of the input articles. Therefore, we use an MLP-based encoder, which is the same as our model, to encode the full length of articles. In our preliminary experiments, the MLP-based encoder with full length articles achieves better scores than the RNN/CNN-based encoder with limited length articles. It shows that the seq2seq model gets low scores on all relevant metrics, mainly because of the mode collapse problem as described in Section Challenges. Unlike seq2seq, IR is based on a retrieval framework, so it achieves much better performance.

\subsection{Generative Evaluation} 

Following previous work~\cite{QinEA2018}, we evaluate the models under the generative evaluation setting. The retrieval-based models generate the comments by selecting a comment from the candidate set. The candidate set contains the comments in the training set. Unlike the retrieval evaluation, the reference comments may not appear in the candidate set, which is closer to real-world settings. Generative-based models directly generate comments without a candidate set. We compare the generated comments of either the retrieval-based models or the generative models with the five reference comments. We select four popular metrics in text generation to compare the model outputs with the references: BLEU~\cite{bleu}, METEOR~\cite{meteor}, ROUGE~\cite{rough}, CIDEr~\cite{cider}.

\subsubsection{Result}

Table~\ref{tab_gen} shows the performance for our models and the baselines in generative evaluation. Similar to the retrieval evaluation, our proposed model outperforms the other unsupervised methods, which are TF-IDF, NVDM, and LDA, in generative evaluation. Still, the supervised IR achieves better scores than the seq2seq model. With the help of our proposed model, both IR and S2S achieve an improvement under the semi-supervised scenarios. 

\subsection{Analysis and Discussion}

We analyze the performance of the proposed method under the semi-supervised setting. We train the supervised IR model with different numbers of paired data. Figure~\ref{semi_fig} shows the curve (blue) of the recall\@1 score. As expected, the performance grows as the paired dataset becomes larger. We further combine the supervised IR with our unsupervised model, which is trained with full unpaired data (4.8M) and different number of paired data (from 50K to 4.8M). It shows that IR+Proposed can outperform the supervised IR model given the same paired dataset. It concludes that the proposed model can exploit the unpaired data to further improve the performance of the supervised model.

Although our proposed model can achieve better performance than previous models, there are still remaining two questions: why our model can outperform them, and how to further improve the performance. To address these queries, we perform error analysis to analyze the error types of our model and the baseline models. We select TF-IDF, S2S, and IR as the representative baseline models. We provide 200 unique comments as the candidate sets, which consists of four types of comments as described in the above retrieval evaluation setting: Correct, Plausible, Popular, and Random. We rank the candidate comment set with four models (TF-IDF, S2S, IR, and Proposed+IR), and record the types of top-1 comments.

Figure~\ref{error_fig} shows the percentage of different types of top-1 comments generated by each model. It shows that TF-IDF prefers to rank the plausible comments as the top-1 comments, mainly because it matches articles with the comments based on the similarity of the lexicon. Therefore, the plausible comments, which are more similar in the lexicon, are more likely to achieve higher scores than the correct comments. It also shows that the S2S model is more likely to rank popular comments as the top-1 comments. The reason is the S2S model suffers from the mode collapse problem and data sparsity, so it prefers short and general comments like ``Great'' or ``That's right'', which appear frequently in the training set. The correct comments often contain new information and different language models from the training set, so they do not obtain a high score from S2S.

IR achieves better performance than TF-IDF and S2S. However, it still suffers from the discrimination between the plausible comments and correct comments. This is mainly because IR does not explicitly model the underlying topics. Therefore, the correct comments which are more relevant in topic with the articles get lower scores than the plausible comments which are more literally relevant with the articles. With the help of our proposed model, proposed+IR achieves the best performance, and achieves a better accuracy to discriminate the plausible comments and the correct comments. Our proposed model incorporates the topic information, so the correct comments which are more similar to the articles in topic obtain higher scores than the other types of comments. According to the analysis of the error types of our model, we  still need to focus on avoiding predicting the plausible comments.

\section{Related Work}

\subsection{Article Comment}

There are few studies regarding machine commenting. Qin et al.~\shortcite{QinEA2018} is the first to propose the article commenting task and a dataset, which is used to evaluate our model in this work. More studies about the comments aim to automatically evaluate the quality of the comments. Park et al.~\shortcite{ParkSDE16} propose a system called CommentIQ, which assist the comment moderators in identifying high quality comments. Napoles et al.~\shortcite{NapolesTPRP17} propose to discriminating engaging, respectful, and informative conversations. They present a Yahoo news comment threads dataset and annotation
scheme for the new task of identifying ``good'' online conversations. 
More recently, Kolhaatkar and Taboada~\shortcite{KolhatkarT17} propose a model to classify the comments into constructive comments and non-constructive comments. In this work, we are also inspired by the recent related work of natural language generation models\cite{DBLP:journals/corr/abs-1803-01465,DBLP:journals/corr/abs-1805-05181}.

\subsection{Topic Model and Variational Auto-Encoder}

Topic models~\cite{Blei12} are among the most widely used models for learning unsupervised representations of text. One of the most popular approaches for modeling the topics of the documents is the Latent Dirichlet Allocation~\cite{Blei03}, which assumes a discrete mixture distribution over topics is sampled from a Dirichlet prior shared by all documents. In order to explore the space of different modeling assumptions, some black-box inference methods~\cite{Mnih14,Ranganath14} are proposed and applied to the topic models.

Kingma and Welling~\shortcite{vae} propose the Variational Auto-Encoder (VAE) where the generative model and the variational posterior are based on neural networks. VAE has recently been applied to modeling the representation and the topic of the documents. Miao et al.~\shortcite{NVDM} model the representation of the document with a VAE-based approach called the Neural Variational Document Model (NVDM). However, the representation of NVDM is a vector generated from a Gaussian distribution, so it is not very interpretable unlike the multinomial mixture in the standard LDA model. To address this issue, Srivastava and Sutton~\shortcite{nvlda} propose the NVLDA model that replaces the Gaussian prior with the Logistic Normal distribution to approximate the Dirichlet prior and bring the document vector into the multinomial space. More recently, Nallapati et al.~\shortcite{sengen} present a variational auto-encoder approach which models the posterior over the topic assignments to sentences using an RNN.

\section{Conclusion}

We explore a novel way to train a machine commenting model in an unsupervised manner. According to the properties of the task, we propose using the topics to bridge the semantic gap between articles and comments. We introduce a variation topic model to represent the topics, and match the articles and comments by the similarity of their topics. Experiments show that our topic-based approach significantly outperforms previous lexicon-based models. The model can also profit from paired corpora and achieves state-of-the-art performance under semi-supervised scenarios.

\bibliographystyle{aaai}
\bibliography{aaai2018}

\begin{thebibliography}{}

\bibitem[\protect\citeauthoryear{Auvolat and Vincent}{2015}]{mips1}
Auvolat, A., and Vincent, P.
\newblock 2015.
\newblock Clustering is efficient for approximate maximum inner product search.
\newblock {\em CoRR} abs/1507.05910.

\bibitem[\protect\citeauthoryear{Bahdanau, Cho, and Bengio}{2014}]{attention}
Bahdanau, D.; Cho, K.; and Bengio, Y.
\newblock 2014.
\newblock Neural machine translation by jointly learning to align and
  translate.
\newblock {\em CoRR} abs/1409.0473.

\bibitem[\protect\citeauthoryear{Banerjee and Lavie}{2005}]{meteor}
Banerjee, S., and Lavie, A.
\newblock 2005.
\newblock {METEOR:} an automatic metric for {MT} evaluation with improved
  correlation with human judgments.
\newblock In {\em Proceedings of the Workshop on Intrinsic and Extrinsic
  Evaluation Measures for Machine Translation and/or Summarization@ACL 2005,
  Ann Arbor, Michigan, USA, June 29, 2005},  65--72.

\bibitem[\protect\citeauthoryear{Blei, Ng, and Jordan}{2003}]{Blei03}
Blei, D.~M.; Ng, A.~Y.; and Jordan, M.~I.
\newblock 2003.
\newblock Latent dirichlet allocation.
\newblock {\em Journal of Machine Learning Research} 3:993--1022.

\bibitem[\protect\citeauthoryear{Blei}{2012}]{Blei12}
Blei, D.~M.
\newblock 2012.
\newblock Probabilistic topic models.
\newblock {\em Commun. {ACM}} 55(4):77--84.

\bibitem[\protect\citeauthoryear{Guo \bgroup et al\mbox.\egroup }{2016}]{mips2}
Guo, R.; Kumar, S.; Choromanski, K.; and Simcha, D.
\newblock 2016.
\newblock Quantization based fast inner product search.
\newblock In {\em Proceedings of the 19th International Conference on
  Artificial Intelligence and Statistics, {AISTATS} 2016},  482--490.

\bibitem[\protect\citeauthoryear{Henderson \bgroup et al\mbox.\egroup
  }{2017}]{smartreply}
Henderson, M.; Al{-}Rfou, R.; Strope, B.; Sung, Y.; Luk{\'{a}}cs, L.; Guo, R.;
  Kumar, S.; Miklos, B.; and Kurzweil, R.
\newblock 2017.
\newblock Efficient natural language response suggestion for smart reply.
\newblock {\em CoRR} abs/1705.00652.

\bibitem[\protect\citeauthoryear{Hennig \bgroup et al\mbox.\egroup
  }{2012}]{laplace}
Hennig, P.; Stern, D.~H.; Herbrich, R.; and Graepel, T.
\newblock 2012.
\newblock Kernel topic models.
\newblock In {\em Proceedings of the Fifteenth International Conference on
  Artificial Intelligence and Statistics, {AISTATS} 2012, La Palma, Canary
  Islands, Spain, April 21-23, 2012},  511--519.

\bibitem[\protect\citeauthoryear{Ioffe and Szegedy}{2015}]{batchnorm}
Ioffe, S., and Szegedy, C.
\newblock 2015.
\newblock Batch normalization: Accelerating deep network training by reducing
  internal covariate shift.
\newblock In {\em Proceedings of the 32nd International Conference on Machine
  Learning, {ICML} 2015, Lille, France, 6-11 July 2015},  448--456.

\bibitem[\protect\citeauthoryear{Kingma and Ba}{2014}]{KingmaBa2014}
Kingma, D.~P., and Ba, J.
\newblock 2014.
\newblock Adam: {A} method for stochastic optimization.
\newblock {\em CoRR} abs/1412.6980.

\bibitem[\protect\citeauthoryear{Kingma and Welling}{2013}]{vae}
Kingma, D.~P., and Welling, M.
\newblock 2013.
\newblock Auto-encoding variational bayes.
\newblock {\em CoRR} abs/1312.6114.

\bibitem[\protect\citeauthoryear{Kolhatkar and Taboada}{2017}]{KolhatkarT17}
Kolhatkar, V., and Taboada, M.
\newblock 2017.
\newblock Using new york times picks to identify constructive comments.
\newblock In {\em Proceedings of the 2017 Workshop: Natural Language Processing
  meets Journalism, NLPmJ@EMNLP, Copenhagen, Denmark, September 7, 2017},
  100--105.

\bibitem[\protect\citeauthoryear{Lin and Hovy}{2003}]{rough}
Lin, C., and Hovy, E.~H.
\newblock 2003.
\newblock Automatic evaluation of summaries using n-gram co-occurrence
  statistics.
\newblock In {\em Human Language Technology Conference of the North American
  Chapter of the Association for Computational Linguistics, {HLT-NAACL} 2003}.

\bibitem[\protect\citeauthoryear{Ma \bgroup et al\mbox.\egroup
  }{2018}]{DBLP:journals/corr/abs-1803-01465}
Ma, S.; Sun, X.; Li, W.; Li, S.; Li, W.; and Ren, X.
\newblock 2018.
\newblock Query and output: Generating words by querying distributed word
  representations for paraphrase generation.
\newblock In {\em {NAACL-HLT} 2018},  196--206.

\bibitem[\protect\citeauthoryear{Miao, Yu, and Blunsom}{2016}]{NVDM}
Miao, Y.; Yu, L.; and Blunsom, P.
\newblock 2016.
\newblock Neural variational inference for text processing.
\newblock In {\em Proceedings of the 33nd International Conference on Machine
  Learning, {ICML} 2016},  1727--1736.

\bibitem[\protect\citeauthoryear{Mikolov \bgroup et al\mbox.\egroup
  }{2013}]{MikolovEA2013}
Mikolov, T.; Sutskever, I.; Chen, K.; Corrado, G.~S.; and Dean, J.
\newblock 2013.
\newblock Distributed representations of words and phrases and their
  compositionality.
\newblock In {\em Advances in Neural Information Processing Systems 26: 27th
  Annual Conference on Neural Information Processing Systems 2013.},
  3111--3119.

\bibitem[\protect\citeauthoryear{Mnih and Gregor}{2014}]{Mnih14}
Mnih, A., and Gregor, K.
\newblock 2014.
\newblock Neural variational inference and learning in belief networks.
\newblock In {\em Proceedings of the 31th International Conference on Machine
  Learning, {ICML} 2014, Beijing, China, 21-26 June 2014},  1791--1799.

\bibitem[\protect\citeauthoryear{Nallapati \bgroup et al\mbox.\egroup
  }{2017}]{sengen}
Nallapati, R.; Melnyk, I.; Kumar, A.; and Zhou, B.
\newblock 2017.
\newblock Sengen: Sentence generating neural variational topic model.
\newblock {\em CoRR} abs/1708.00308.

\bibitem[\protect\citeauthoryear{Napoles \bgroup et al\mbox.\egroup
  }{2017}]{NapolesTPRP17}
Napoles, C.; Tetreault, J.~R.; Pappu, A.; Rosato, E.; and Provenzale, B.
\newblock 2017.
\newblock Finding good conversations online: The yahoo news annotated comments
  corpus.
\newblock In {\em Proceedings of the 11th Linguistic Annotation Workshop,
  LAW@EACL 2017, Valencia, Spain, April 3, 2017},  13--23.

\bibitem[\protect\citeauthoryear{Papineni \bgroup et al\mbox.\egroup
  }{2002}]{bleu}
Papineni, K.; Roukos, S.; Ward, T.; and Zhu, W.
\newblock 2002.
\newblock Bleu: a method for automatic evaluation of machine translation.
\newblock In {\em Proceedings of the 40th Annual Meeting of the Association for
  Computational Linguistics},  311--318.

\bibitem[\protect\citeauthoryear{Park \bgroup et al\mbox.\egroup
  }{2016}]{ParkSDE16}
Park, D.~G.; Sachar, S.~S.; Diakopoulos, N.; and Elmqvist, N.
\newblock 2016.
\newblock Supporting comment moderators in identifying high quality online news
  comments.
\newblock In {\em Proceedings of the 2016 {CHI} Conference on Human Factors in
  Computing Systems, San Jose, CA, USA, May 7-12, 2016},  1114--1125.

\bibitem[\protect\citeauthoryear{Qin \bgroup et al\mbox.\egroup
  }{2018}]{QinEA2018}
Qin, L.; Liu, L.; Bi, W.; Wang, Y.; Liu, X.; Hu, Z.; Zhao, H.; and Shi, S.
\newblock 2018.
\newblock Automatic article commenting: the task and dataset.
\newblock In {\em Proceedings of the 56th Annual Meeting of the Association for
  Computational Linguistics, {ACL} 2018},  151--156.

\bibitem[\protect\citeauthoryear{Ranganath, Gerrish, and
  Blei}{2014}]{Ranganath14}
Ranganath, R.; Gerrish, S.; and Blei, D.~M.
\newblock 2014.
\newblock Black box variational inference.
\newblock In {\em Proceedings of the Seventeenth International Conference on
  Artificial Intelligence and Statistics, {AISTATS} 2014, Reykjavik, Iceland,
  April 22-25, 2014},  814--822.

\bibitem[\protect\citeauthoryear{Shen \bgroup et al\mbox.\egroup
  }{2015}]{mips3}
Shen, F.; Liu, W.; Zhang, S.; Yang, Y.; and Shen, H.~T.
\newblock 2015.
\newblock Learning binary codes for maximum inner product search.
\newblock In {\em 2015 {IEEE} International Conference on Computer Vision,
  {ICCV} 2015, Santiago, Chile, December 7-13, 2015},  4148--4156.

\bibitem[\protect\citeauthoryear{Shrivastava and Li}{2014}]{mips4}
Shrivastava, A., and Li, P.
\newblock 2014.
\newblock Asymmetric {LSH} {(ALSH)} for sublinear time maximum inner product
  search {(MIPS)}.
\newblock In {\em Advances in Neural Information Processing Systems 27: Annual
  Conference on Neural Information Processing Systems 2014},  2321--2329.

\bibitem[\protect\citeauthoryear{Srivastava and Sutton}{2017}]{nvlda}
Srivastava, A., and Sutton, C.
\newblock 2017.
\newblock Autoencoding variational inference for topic models.
\newblock In {\em ICLR 2017}.

\bibitem[\protect\citeauthoryear{Sutskever, Vinyals, and Le}{2014}]{seq2seq}
Sutskever, I.; Vinyals, O.; and Le, Q.~V.
\newblock 2014.
\newblock Sequence to sequence learning with neural networks.
\newblock In {\em Advances in Neural Information Processing Systems 27: Annual
  Conference on Neural Information Processing Systems 2014},  3104--3112.

\bibitem[\protect\citeauthoryear{Vedantam, Zitnick, and Parikh}{2015}]{cider}
Vedantam, R.; Zitnick, C.~L.; and Parikh, D.
\newblock 2015.
\newblock Cider: Consensus-based image description evaluation.
\newblock In {\em {IEEE} Conference on Computer Vision and Pattern Recognition,
  {CVPR} 2015},  4566--4575.

\bibitem[\protect\citeauthoryear{Xu \bgroup et al\mbox.\egroup
  }{2018}]{DBLP:journals/corr/abs-1805-05181}
Xu, J.; Sun, X.; Zeng, Q.; Ren, X.; Zhang, X.; Wang, H.; and Li, W.
\newblock 2018.
\newblock Unpaired sentiment-to-sentiment translation: {A} cycled reinforcement
  learning approach.
\newblock In {\em {ACL} 2018},  979--988.

\end{thebibliography}

\end{document}